\documentclass{llncs}
\usepackage{graphicx}
\usepackage{bigdelim}
\usepackage{multirow}
\usepackage{amsfonts}
\usepackage{framed}
\usepackage{pifont}
\usepackage{wrapfig}
\usepackage{color}
\usepackage{subfigure}
\usepackage{times}
\usepackage[right]{eurosym}
\usepackage{xspace}
\usepackage{todonotes}
\usepackage{listings} % Thao: Algorithmen
\usepackage{eurosym}
\usepackage{breakurl}
\usepackage[hyphens]{url}
\usepackage{tabularx}

\usepackage[defblank]{paralist}
\usepackage{framed}
\definecolor{shadecolor}{RGB}{217,229,242}

\begin{document}
%\shorthandoff{"}
%
%\frontmatter          % for the preliminaries
%
%\pagestyle{headings}  % switches on printing of running heads
%\addtocmark{Mining Staff Assignment Rules} % additional mark in the TOC
%
% }}}
\newcommand{\None}{\raisebox{-3pt}{\includegraphics[height=0.17in]{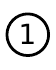}}}% {{{
\newcommand{\Ntwo}{\raisebox{-3pt}{\includegraphics[height=0.17in]{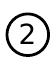}}}
\newcommand{\Nthree}{\raisebox{-3pt}{\includegraphics[height=0.17in]{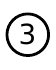}}}
\newcommand{\Nfour}{\raisebox{-3pt}{\includegraphics[height=0.17in]{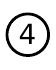}}}
\newcommand{\Nfive}{\raisebox{-3pt}{\includegraphics[height=0.17in]{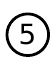}}}
\newcommand{\Nsix}{\raisebox{-3pt}{\includegraphics[height=0.17in]{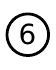}}}
\newcommand{\Nseven}{\raisebox{-3pt}{\includegraphics[height=0.17in]{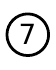}}}
\newcommand{\Neight}{\raisebox{-3pt}{\includegraphics[height=0.17in]{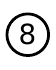}}}
\newcommand{\Nnine}{\raisebox{-3pt}{\includegraphics[height=0.17in]{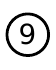}}}
\newcommand{\Nten}{\raisebox{-3pt}{\includegraphics[height=0.17in]{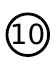}}}
\newcommand{\Neleven}{\raisebox{-3pt}{\includegraphics[height=0.17in]{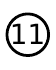}}}

\begin{titlepage}

\title{The NNN Formalization: Review and Development of Guideline Specification in the Care Domain}
\subtitle{Technical Report}

\author{Georg Kaes\inst{1} \and J\"urgen Mangler\inst{1} \and Stefanie Rinderle-Ma\inst{1} \and \\ Ralph Vigne\inst{2}}
\institute{University of Vienna, Faculty of Computer Science, Austria\\
CERN\\
\email{\{georg.kaes, juergen.mangler, stefanie.rinderle-ma\}@univie.ac.at}
\email{ralph.vigne@cern.ch}}

\end{titlepage}
\maketitle

% }}}

\section{Introduction}
\label{ref:intro}

This technical report describes a formalization for nursing knowledge found in the NANDA, NIC and NOC (NNN for short) standards.

Data about nursing diagnoses and treatments consists of the following three {\bf NNN} knowledge sources that are widely accepted and used in the field of care \cite{nanda-i_2012}, i.e., \textit{NANDA}\footnote{\underline{N}orth \underline{A}merican \underline{N}ursing \underline{D}iagnosis \underline{A}ssociation, \url{www.nanda.org}} containing all basic information about the diagnoses, the \textit{Nursing Interventions Classification (NIC)} containing all treatments which can be executed for improving the patients condition, and the \textit{Nursing Outcomes Classification (NOC)} containing all nursing outcomes which can be reached after a therapy.

Based on the NNN knowledge, our overall goal is to enable the guideline-driven process development and adaptation in the care domain. 
Within the ACaPlan\footnote{\underline{A}daptive \underline{Ca}re \underline{Plan}ning: \burl{http://cs.univie.ac.at/research/projects/projekt/infproj/1033/}} project, we work closely together with experts from the care domain. 
This paper contributes a first step towards this goal by providing a formalization method for the NNN knowledge sources in such a way that this information can be directly utilized for creation and adaptation of individual patient treatment processes. 

The central aspects of designing a corresponding formalization are as follows:

Methodologically, we first analyze which information of the NNN knowledge sources has to be included in the formalization based on studying NNN documentation and discussions with experts from the care domain resulting in the NNN taxonomy (contribution 1). 
Then we will evaluate existing standards, primarily from the medical domain, i.e., GLIF \cite{boxwala2004glif3}, Asbru \cite{kosara1998asbruview}, and ARDEN Syntax \cite{Pryor1993}, with respect to their support for the NNN taxonomy, illustrated by the use case FATIGUE. 
The resulting evaluation report and open issues (contribution 2) will serve as input for the design of the NNN formalization (contribution 3). 
The NNN formalization will be evaluated based on use case FATIGUE, results from discussing with domain experts, and possible application in other domains. 

In detail, NNN knowledge sources are introduced and analyzed in Section \ref{ref:careplanningnanda}. 
Existing standards from the medical domain are evaluated in Section \ref{ref:evaluation}. 
Section \ref{ref:nandaformalization} presents the NNN formalization. 
Section \ref{ref:discussion} focuses on lessons learned and a discussion about further advantages which arise from our contribution.

\section{NANDA, NIC, and NOC Knowledge Sources}
\label{ref:careplanningnanda}

The goal of NANDA, NIC and NOC is the development and unification of nursing diagnoses that are the basis for all processes related to care, as they contribute a consistent terminology and ease the phrasing and documentation \cite{marlies_ehmann_pflegediagnosen_2004}.
NANDA contains 206 nursing diagnoses \cite{_nanda_2012}, which are available for care attendants in various text documents. 
Based on the nursing diagnoses defined in NNN, nurses can determine a patients state and in further steps are able to create therapies and define intended outcomes, which are used for treating the observed symptoms.

For getting an overview over the different sections and contents of a diagnosis we studied a variety of these documents (\cite{marilynn_e._doenges_pflegediagnosen_2002,marlies_ehmann_pflegediagnosen_2004,_nanda_example_1_2012,_nanda_example_2_2012,_nanda_glossary_2012}) and conferred with domain experts who showed us the most relevant parts of the given taxonomy. 

\begin{figure}
  \centering
  \includegraphics[width=0.8\textwidth]{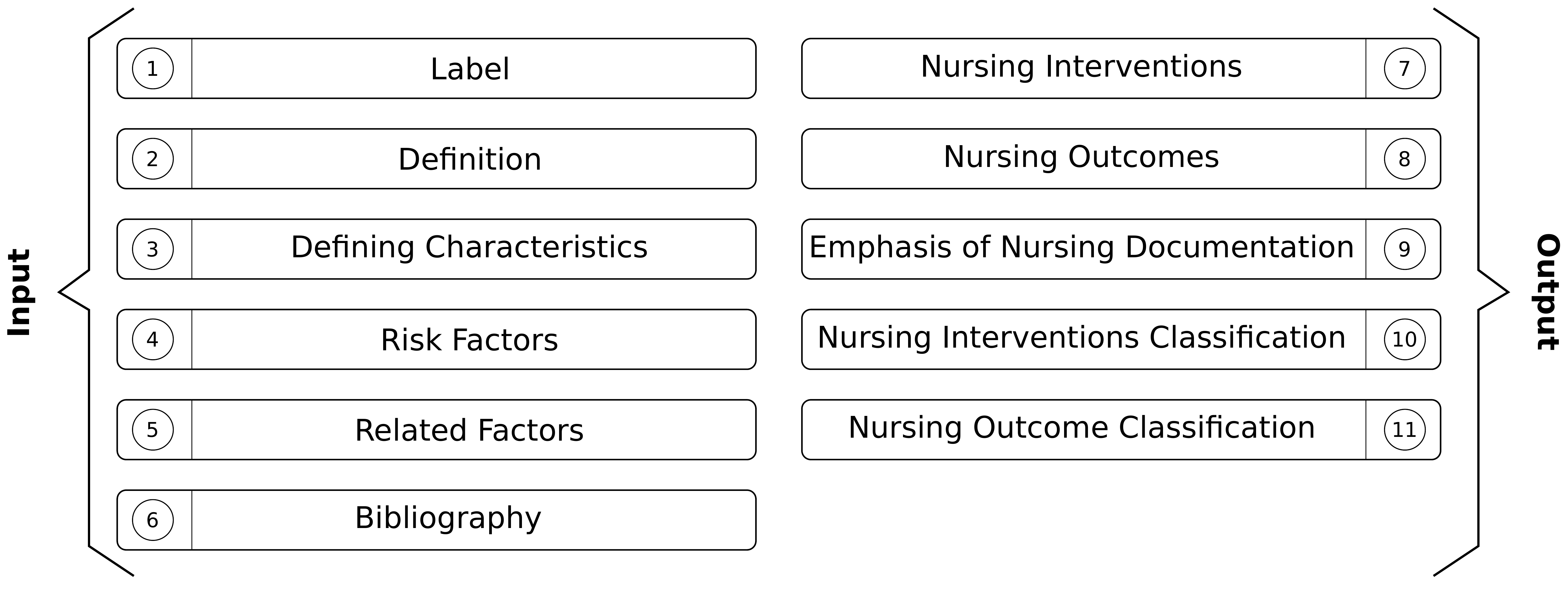}
  \caption{Building Blocks for NNN Formalization}
  \label{fig:nanda:infos}
\end{figure}

As a fist step, we analyzed and aggregated the necessary building blocks of NNN, which are depicted in Figure \ref{fig:nanda:infos}. 
In the following, these building blocks are shortly described and illustrated by the use case FATIGUE. 

\begin{itemize}

  \item \None,  \Ntwo:  provide identification of the diagnosis, e.g., FATIGUE, and describes the most important characteristics of the given diagnosis in one or more natural language sentences respectively.
      For FATIGUE, the descripion reads as follows:
      ``An overwhelming sustained sense of exhaustion and decreased capacity for physical and mental work at usual level'' \cite{_nanda_example_2_2012}.
 
  \item \Nthree: are the symptoms which imply the fact that there is a hardship in the patients circumstances. 
      Those characteristics, which can be either subjectively (``I feel a bit dizzy today.'') or objectively (``Your nose is bleeding.'') determined, are further informational assets for the care attendant to review or approve the diagnosis \cite{marilynn_e._doenges_pflegediagnosen_2002}.

  \item \Nfour: are possible causes, which may lead to the given nursing diagnosis \cite{marilynn_e._doenges_pflegediagnosen_2002}. 
      It is a very important fact that care attendants should not concentrate on treating the symptoms a patient shows, but the causes which are the reasons for the current circumstances. 
      For FATIGUE, there are several potential risk factors defined, for example psychological causes such as stress, fear, or depression \cite{marilynn_e._doenges_pflegediagnosen_2002}.

  \item \Nfive: describe risk factors, defining characteristics for possible problems which may lead to future hardships for the patient. 
      Diagnoses which contain these kind of factors only concentrate on risks, and therefore do not implement any defining characteristics (as there are none yet).

  \item \Nsix: contains all sources which led to the development of the given taxonomy. 
      They have to be linked to the appropriate part of the diagnosis.  

\end{itemize}

After defining all information sources related to the current state of the patient, the second class of information refers to the possible treatments, their outcome, and their documentation:

\begin{itemize}

  \item \Nseven: are the concrete tasks which need to be executed for reaching the desired nursing goals given the current circumstances. 
      They have to be defined in a standardized terminology like the Nursing Interventions Classification, or supplemented with resources.

  \item \Neight: define the desired outcomes after the execution of the therapy. 
      As with Nursing Interventions, they either have to be defined within a standardized terminology like the Nursing Outcomes Classification, or supplemented with sources.

  \item \Nnine: are not necessarily part of the NIC taxonomy itself, but nonetheless an important part of a complete definition of a given diagnosis. 
      They are listed explicitly in specialized literature such as \cite{marilynn_e._doenges_pflegediagnosen_2002}.
      The emphases of nursing documentation define the most important facts for the documentation of the executed tasks. 
      Based on the necessity of a situational and regular evaluation of the performed actions, the possibility to totally reconstruct the methods of the acting care attendants is very important. 
      For the diagnosis of FATIGUE there exist several categories, which contain the necessary documentational elements.  

  \item \Nten and \Neleven: list the relevant categories from the Nursing Interventions classification respectively the Nursing Outcomes Classification.

\end{itemize}

\section{Evaluation of Existing Standards: GLIF, Asbru, ARDEN Syntax}
\label{ref:evaluation}

Similarly to NNN knowledge supporting the work of care personnel, Clinical Practice Guidelines (CPGs) define how medical staff has to act in certain situations and are an essential part of modern medicine. 
During the last decades several approaches have been developed aiming at formalizing this medical knowledge in a manner which is also accessible to computers. 
The goal of these Computer-Interpretable Guidelines (CIGs) \cite{Mulyar2007} which are a formal representation of the CPGs, is helping the doctors with their decisions in the best way possible. 

In the following we evaluate three well known approaches - namely the Arden Syntax, Asbru, and GLIF as representatives for CIG standards. 
We chose three well known approaches which we have found in specialized literature \cite{de2008computer} \cite{peleg2003comparing} very often. 
Of course, there do exist more approaches, like the \textit{Evicare} project, which focuses on ``providing evidence-based medicine at the point of care'' \cite{_evicare_2009}, thus increasing quality of patient care. 
Their guideline formalizations are based on the \textit{DeGeL} framework, whose model supports elements common to clinical guidelines. 
\cite{dojat_degel_2003} shows how guidelines defined in Asbru and GEM can be implemented in DeGeL.

In this section, we describe the design of the Arden Syntax, Asbru and GLIF and try a formalization of NNN in the particular approach based on the example of FATIGUE. 
The result will be a discussion of limitations of existing CIG standards with respect to formalizing care-related diagnoses.

\subsection{The Arden Syntax}% {{{
\label{ref:evaluation:arden} 

The Arden Syntax is a guideline formalism implementing a language close to Pascal with the goal of formalizing medical knowledge. 
The basic elements are called Medical Logic Moduls (MLMS), which can be reused in several applications \cite{hripcsak_arden_1990,de_clercq_computer-interpretable_2008}. 
Representing knowledge in distinct, separate modules facilitates the implementation of contents relevant to the respective institution into their own \textit{Electronic Patient Record}-System.

\noindent{\sl Design Principle:} Medical Logic Modules are the basic elements of the Arden Syntax. 
They contain the structure how knowledge is represented and give the medical professionals - if they are implemented in a clinical information system - informations about a patient's condition by using alerts \cite{Pryor1993}.

MLMs consist of the three section {\sl Maintenance}, {\sl Library}, and {\sl Knowledge}. 
The latter contains the actual medical knowledge encoded within different slots, i.e., key-value pairs where the value can be either text, a coded value, or structured data. 
The following MLM excerpt for section Knowledge implementing NANDA diagnosis FATIGUE demonstrates its basic structure (note that some simplifications were made). 

\lstset{caption={Component: Knowledge}}
\begin{tt}
\begin{scriptsize}
\begin{lstlisting}

type: data-driven;;
data:
listless := READ {select listlessness from results where it occurred within the past 1 week};
features := READ {select feature from results where TirednessIndicator = true};
priority: 42;
evoke: ANY OF (listless, features)
logic: 
IF features='very tired' OR listless > 84 conclude
action:
WRITE 'Assess the patient's ability to perform activities of daily living' TO nurse_infoscreen

\end{lstlisting}
\end{scriptsize}
\end{tt}

The knowledge component contains the medical knowledge. 
In the data slot we have defined two variables. 
The first one saves a fictive indicator for listlessness, which is going to be read from a database, the second one saves a corresponding value in the key \textit{features}. 
The MLMs priority is set to 42, and the MLM will be, as defined in the evoke slot, evoked as soon as one of those two variables is set. 
The logic slot defines the rule, which has to be evaluated for executing the action slot, if it returns true. 
In this example, the evoke slot returns true, if the feature is ``very tired'', or the value on the listlessness scale is bigger than 84. 
The action defines to write an approriate message to an info screen (named \textit{nurse\_infoscreen}). 

According to the building block defined in Fig. \ref{fig:nanda:infos} we summarize as follows:

\begin{itemize}

  \item \None is best placed into the \textit{title slot}.

  \item \Ntwo can be defined in the \textit{explanation key} of the \textit{Library} section.

  \item \Nthree can be considered as part of \textit{evoke} $\rightarrow$ \textit{keywords slot}

  \item \Nfour, \Nfive, \Nten and \Neleven are not supported at all.
 
  \item \Nsix has its counterpart in the \textit{citations slot}.

  \item \Nseven and \Nnine are matching with \textit{action slots}. 
      Although the intention is similar it should be noted that there is one major conceptual difference: 
      the \textit{all-or-nothing} aproach of ARDEN. 
      This makes NNN definitions very complex as each separate \Nseven spawns a new MLM.

  \item \Neight, although it is similar to the \textit{library} $\rightarrow$ \textit{purpose slot} the difference is still to big to be neglected

\end{itemize}
% }}}

\subsection{Asbru}% {{{
\label{nanda:cig:asbru}

Asbru is a modeling notation which emphasizes the temporal structures of medical plans \cite{Shahar1998}. 
The underlying \textit{skeletal plan} leaves room for the temporal planning of the individual actions, which increases flexibility. 
This way, plans can be adapted to changing circumstances, for example if medical professionals are needed on other stations. 
Asbru consists of 2 phases. While during design phase, you can define timely confined actions and alternatives as reactions for conditions of any patient, during execution phase these resulting skeleton planes get instantiated for a concrete patient.
This emphasizes the generation of practical plans. 
The language, which is based on XML and defined over a DTD shows a syntax close to Lisp. 

\noindent{\sl Design Principles:} Plans in Asbru consist of several elements such as {\sl Preferences}, {\sl Intentions}, and {\sl Plan Body}. 
Plan hierarchies can be also defined \cite{de_clercq_computer-interpretable_2008}.
With \textit{AsbruView} a visualization approach for plans has been proposed \cite{kosara1998asbruview}. 

For demonstrating the XML structure of Asbru, selected parts of the NNN diagnosis FATIGUE have been implemented in this notation. 
The following listing shows the implementation of an exemplary section of the {\sl Intentions} part. 
The desired nursing outcome of this plan is to increase the energy level (a fictive scale for measuring the energy a patient currently has) of a patient to a value of at least 50 over 3 days.

\lstset{caption={Intentions}}
\begin{tt}
\begin{scriptsize}
\begin{lstlisting}
<intentions>
    <intention label="PowerEnergy" type="overall-state " verb="achieve" importance="1">
        <temporal-pattern>
            <parameter-proposition parameter-name="patientEnergy">
                <value-description type="greater-than">
                    <numerical-constant unit="E" value="50"/>
                </value-description>
                <context><context-ref name="patientEnergy"/></context>
                <time-annotation>
                    <time-range>
                        <duration>
                            <minimum>
                                <numerical-constant unit="d" value="3"/>
                            </minimum>
                        </duration>
                    </time-range>
                    <self/>
                </time-annotation>
            </parameter-proposition>
        </temporal-pattern>
    </intention>
</intentions>
\end{lstlisting}
\end{scriptsize}
\end{tt}

Although Asbru provides a flexible scheme for formalizing medical guidelines, it can only be of limited use to care related guidelines. 
In the following we list the main differnces according to the building blocks defined in Fig. \ref{fig:nanda:infos}: 

\begin{itemize}

  \item \None can be defined in the \textit{title attribute} of the plan
itself. 

  \item \Nthree can be defined in the \textit{preconditions}, but it should be
noted theat doing so leads to the loss of support for precautions.

  \item \Nsix could be put into the \textit{comment} tag which is applicable to
a lot of tags.

  \item \Nseven and \Nnine are going to be defined as nested sub plans by using
several \textit{plan-bodys}. 

  \item \Neight match directly with \textit{intentions} wich additionally act
as a container for \Neleven.

  \item \Ntwo, \Nfour, \Nfive and \Nten can not be defined (leaving a
misusing of the \textit{comment tag} aside) 

\end{itemize}
% }}}

\subsection{The GuideLine Interchange Format (GLIF)}% {{{
\label{nanda:cig:glif}

GLIF (current version GLIF3) \cite{boxwala2004glif3} has the goal of easing the interchange of guidelines between different institutions and platforms. 
It offers structures, which make it easy for users to understand the purpose of the guideline, as well as such, which are of use for decision support systems.
In GLIF all classes are represented by UML class diagrams. 
The actual structure of the guidelines is defined in RDF. Other constraints are defined by in OCL \cite{peleg_guideline_2004}. 

\noindent{\sl Design Principles:} GLIF consists of two components: the \textit{GLIF model}, and the \textit{GLIF syntax}. 
While guidelines are implemented as instances of the classes defined in the GLIF model, the representation of the knowledge covered by the guideline is specified in the GLIF syntax. 
\textit{Guideline} is the class which describes a guideline.
All general attributes, like the name, the author, the purpose, selection criteria (which are implemented as criterion objects), steps, the starting point, and further information (didactics) are part of this class. 
The class \textit{Guideline\_Collection} describes a collection of classes which belong together. 
The class \textit{Supplemental Material List} can be used for describing further information about the class, like the documentation. 
The class \textit{Guideline\_Expressions} represents all kinds of expressions, from simple strings like ``weight'' to logical expressions, like ``weight $<$ 90''.
Further on, GLIF proposes a 3-level-model consisting of the conceptual, computable, and the implemetable level. 
All these levels give a different degree of abstraction, and can therefore be useful to different kinds of users (from real persons to computer programs).

For showing the structure of GLIF, again we have implemented selected parts of the NANDA diagnosis FATIGUE. 
In the following we show how the classes \textit{Guideline}, \textit{Action Step} and \textit{Action Specification} are being implemented.

The class \textit{Guideline} contains general information. 
In this example we have defined two criteria which have to evaluate to \textit{true} before starting the guideline; these are \textit{LackOfEnergy} and \textit{NeedForEnergy}. 
The \textit{Steps} defines the steps which are part of the guideline. 
In this case, only the two simple steps  \textit{MedicationWatch} and \textit{DailyLivingWatch} are implemented. 
\textit{First step} defines, that we first have a look on the medication.

\lstset{caption={Class: Guideline}}
\begin{lstlisting}
name: Fatigue
author: Georg Kaes
intention: Capacity to sustain activity
eligibility\_criteria: LackOfEnergy, NeedForEnergy, [...]
didactics: An overwhelming sustained sense of exhaustion and decreased capacity for physical and mental work at usual level
step: MedicationWatch, DailyLivingWatch, ...
first step: MedicationWatch
\end{lstlisting}

The Action Steps contain general information about the actions, like which step follows the current action. 
Our definition implies that the step \textit{DailyLivingWatch} is to be executed after the execution of \textit{MedicationWatch}. 

\lstset{caption={Class: Action Step}}
\begin{lstlisting}
action: AS\_MedicationWatch
subguideline: null
next\_step: DailyLivingWatch

name: DailyLivingWatch
action: AS\_DailyLiving
subguideline: null
next\_step: [...]
\end{lstlisting}

Besides that, the \textit{Action Specification} is defined, where further informations about the step are given.

\lstset{caption={Class: Action Specification}}
\begin{lstlisting}
name: AS\_MedicationWatch
patient\_data: Patient\_Medication\_Overview
description: Evaluate the patient’s routine prescription and over-the-counter medications
didactics: [list of supporting didactic materials]
\end{lstlisting}

\lstset{caption={Class: Action Specification}}
\begin{lstlisting}
name: AS\_DailyLiving
patient\_data: Patient\_DailyLiving\_Overview
description: Assess the patient’s ability to perform activities of daily living
didactics: [list of supporting didactic materials]
\end{lstlisting}

GLIF is a comprehensive approach, which allows medical guidelines to be modeled from various points of views. 
But for implementing NANDA in this formalization, we found that some important features are missing (again we reuse the building blocks defined in Fig. \ref{fig:nanda:infos}):

\begin{itemize}

  \item \None match the \textit{name} property of the
\textit{Guideline} class

  \item \Ntwo is equivalent to the \textit{didactics} property

  \item \Nthree can be expressed as a part \textit{eligibility\_criteria} property while \Neight is
represented by the \textit{intentions} property of the same class

  \item \Nfour and  \Nfive have no match at all

  \item \Nsix can be implemented either as instance of the \textit{Supplemental
Material} class or in the corresponding didactics

  \item \Nseven and \Nnine are potential instances of the \textit{action step}
class

  \item \Nten and \Neleven can be added to the \textit{didactics} property of
the \textit{Action Specification} class

\end{itemize}
% }}}

\subsection{Summarizing the Evaluation Results }% {{{

In this section three representative standards for CIGs have been evaluated based on their capability to implement the NNN building blocks \None to \Neleven as summarized in Figure \ref{fig:nanda:infos}. 
Apparently, none of these approaches can implement the entire structure of the NNN knowledge. 
Particularly, an implementation for risks and related factors of a given diagnosis are missing. 
Hence, in order to meet our requirement to be able to formalize the NNN knowledge in a complete manner, we will introduce the NNN formalization for application in the care domain. 

\begin{table}
\setlength{\tabcolsep}{07pt}
\small
%\centering
\caption{Evaluation of CIG Standards along NNN Building Blocks}
\label{tab:Results}  
\begin{tabular}{|l|c|c|c|c|c|c|}
%\hline
%& \None & \Ntwo & \Nthree & \Nfour & \Nfive & \Nsix & \Nseven & \Neight & \Nnine & \Nten & \Neleven \\\hline
%Arden Syntax & + & + & +  & - & - & + & + & 0 & + & - & - \\\hline
%Asbru & + & - & 0 & - & - & + & + & + & + & - & + \\\hline
%GLIF & + & + & + & - & - & + & + & + & + & 0 & 0 \\\hline
\hline
& \None & \Ntwo & \Nthree & \Nfour & \Nfive & \Nsix  \\\hline
Arden Syntax & + & + & +  & - & - & + \\\hline
Asbru & + & - & 0 & - & - & + \\\hline
GLIF & + & + & + & - & - & + \\\hline
\end{tabular}
\\[10pt]
\begin{tabular}{|l|c|c|c|c|c|}
\hline
& \Nseven & \Neight & \Nnine & \Nten & \Neleven \\\hline
Arden Syntax & + & 0 & + & - & - \\\hline
Asbru & + & + & + & - & + \\\hline
GLIF & + & + & + & 0 & 0 \\\hline
\multicolumn{6}{c}{+: supported, -: not supported, 0: workaround} 
\end{tabular}
\end{table}% }}}

\section{NNN Formalization}
\label{ref:nandaformalization}

The following section describes a way of formalizing NNN. 
After a short introduction into the structure, the different parts are described in detail.

\subsection{Overview}% {{{
\label{nanda:guideline:uebersicht}

The NNN formalization (based on XML) is divided into the following three distinct sections. 

\begin{enumerate}
    \item \textbf{Meta:} Meta-Information about the guideline
    \item \textbf{Custom:} institution-specific preferences for certain tasks
    \item \textbf{Guideline:} information about the guideline itself
\end{enumerate}

Before describing the structure itself, we want to describe some elements that may reoccur throughout the sections:

\begin{itemize}

  \item \textbf{$<$hints$>$} can be added to various elements.
      Each of these elements may consist of one or more $<$hint$>$ elements, defining its origin (\textit{from}) and its purpose (\textit{text}) in its attributes.
      Listing \ref{lst:rng:hints} defines the RNG schema for hints.

  \item \textbf{$<$examples$>$} can be added to support nurses when deciding which treaments to apply.
      Listing \ref{lst:rng:examples} defines the RNG schema for examples.

  \item \textbf{$<$inputs$>$} are used to enforce comprehensive documentation.
      A task may may require more then one parameters of a specific type (e.g. natural numbers, scales, \ldots) for its comprehensive doumentation, e.g. saving the current blood pressure (both systolic and diastolic) of a patient.
      Choosing the right type further enables to reuse the information defined in the NOC when scoring them. 
      To enable this, $<$outcome$>$ elements and $<$task$>$ elements support $<$inputs$>$. 
      To support also complex data structures (e.g. systolic and diastolic combined as blood pressure), $<$inputs$>$ elements may nest multiple $<$input$>$ elements, where each of them contains the attributes \textit{label} describing the meaning of it.
      Listing \ref{lst:rng:inputs} defines the RNG schema for inputs.

  \item Each element containing care relevant information defines a \textbf{score attribute} containing a natural number between 1 and 10. 
      We intend to use this attribute to express preferences to decision support systems (DSS).

\end{itemize}

\lstset{label={lst:rng:hints},caption={RNG schema for the hints element}}% {{{
\begin{tt}
\begin{scriptsize}
\begin{lstlisting}
<define name="refhints">
  <optional>
    <element name="hints">
      <oneOrMore>
        <element name="hint">
          <optional>
            <attribute name="from">
              <text/>
            </attribute>
          </optional>
          <attribute name="text">
            <text/>
          </attribute>
          <optional>
            <attribute name="score">
              <data type="integer"/>
            </attribute>
          </optional>
        </element>
      </oneOrMore>
    </element>
  </optional>
</define>
\end{lstlisting}
\end{scriptsize}
\end{tt}% }}}

\lstset{label={lst:rng:examples},caption={RNG schema for the examples element}}% {{{
\begin{tt}
\begin{scriptsize}
\begin{lstlisting}
<define name="refexamples">
  <optional>
    <element name="examples">
      <oneOrMore>
        <element name="example">
          <attribute name="text">
            <text/>
          </attribute>
          <optional>
            <attribute name="score">
              <data type="integer"/>
            </attribute>
          </optional>
          <optional>
            <ref name="nestedexamples"/>
          </optional>
        </element>
      </oneOrMore>
    </element>
  </optional>
</define>
<define name="nestedexamples">
  <optional>
    <element name="examples">
      <oneOrMore>
        <element name="example">
          <attribute name="text">
            <text/>
          </attribute>
          <optional>
            <attribute name="score">
              <data type="integer"/>
            </attribute>
          </optional>
        </element>
      </oneOrMore>
    </element>
  </optional>
</define>
\end{lstlisting}
\end{scriptsize}
\end{tt}% }}}

\lstset{label={lst:rng:inputs},caption={RNG schema for the inputs element}}% {{{
\begin{tt}
\begin{scriptsize}
\begin{lstlisting}
<define name="refinputs">
  <optional>
    <element name="inputs">
      <oneOrMore>
        <element name="input" ns="http://relaxng.org/ns/structure/1.0">
          <attribute name="label">
            <text/>
          </attribute>
          <ref name="any"/>
        </element>
      </oneOrMore>
    </element>
  </optional>
</define>
<define name="any">
  <element>
    <anyName/>
    <zeroOrMore>
      <choice>
        <attribute>
          <anyName/>
        </attribute>
        <text/>
        <ref name="any"/>
      </choice>
    </zeroOrMore>
  </element>
</define>
\end{lstlisting}
\end{scriptsize}
\end{tt}% }}}

Additionally to the RNG schemes, examples based on the case study of nursing diagnosis FATIGUE demonstrate several parts of our formalization. 
The formalized nursing knowledge can be found in NANDA diagnosis repositories like \cite{factors_fatigue} and \cite{marilynn_e._doenges_pflegediagnosen_2002}. 
This example only serves the puropse of demonstrating the comprehensiveness of our formalization - for real life application, nursing professionals have to implement a practical formalization including relevant scales for a full documentation. 
This care-domain specific knowledge is out of the scope of this technical report.
% }}}

\subsection{Meta}% {{{
\label{nanda:guideline:meta}

This section contains general information about the guidline. 
It therefore includes the bulding blocks \None, \Ntwo and \Nsix (see Fig. \ref{fig:nanda:infos}). 
Information about the \textit{state} (i.e research, implementing, testing, running or expired define by \cite{Pryor1993}), the \textit{author}, the \textit{validator}, the \textit{implementer}, and various \textit{dates} of the guidline are also part of this section.

Listing \ref{lst:rng:meta} shows the RNG schema for the meta section, and listing \ref{lst:fatigue:meta} implements the meta section for FATIGUE exemplarily.

\lstset{label={lst:rng:meta},caption={RNG schema for the Meta section}}
\begin{tt}
\begin{scriptsize}
\begin{lstlisting}
<element name="meta">
  <element name="title">
    <attribute name="text">
      <text/>
    </attribute>
  </element>
  <element name="definition">
    <attribute name="text">
      <text/>
    </attribute>
    <attribute name="theme">
      <text/>
    </attribute>
    <ref name="refhints"/>
  </element>
  <element name="version">
    <attribute name="id">
      <text/>
    </attribute>
  </element>
  <element name="validation">
    <attribute name="status">
      <choice>
        <value>research</value>
        <value>implementing</value>
        <value>testing</value>
        <value>running</value>
        <value>expired</value>
      </choice>
    </attribute>
  </element>
  <optional>
    <element name="institution">
      <attribute name="name">
        <text/>
      </attribute>
    </element>
  </optional>
  <optional>
    <element name="author">
      <attribute name="name">
        <text/>
      </attribute>
    </element>
  </optional>
  <optional>
    <element name="validator">
      <attribute name="name">
        <text/>
      </attribute>
    </element>
  </optional>
  <optional>
    <element name="implementer">
      <attribute name="name">
        <text/>
      </attribute>
    </element>
  </optional>
  <element name="date">
    <attribute name="text">
      <data type="date"/>
    </attribute>
  </element>
</element>
\end{lstlisting}
\end{scriptsize}
\end{tt}

\lstset{label={lst:fatigue:meta},caption={The META section for FATIGUE}}
\begin{tt}
\begin{scriptsize}
\begin{lstlisting}
<meta>
    <title text="fatigue"/>
    <definition text="An overwhelming, sustained sense of exhaustion and decreased capacity for physical and mental work at usual level"/>
    <version id="1.0"/>
    <validation status="implementing"/>
    <institution name="University of Vienna"/>
    <author name="Georg Kaes"/>
    <validator name="Stefanie Rinderle-Ma"/>
    <implementer name="Juergen Mangler"/>
    <date text="2013-04-01"/>
</meta>
\end{lstlisting}
\end{scriptsize}
\end{tt}
% }}}

\subsection{Custom}% {{{

The \textit{custom} section contains institution specific preferences regarding $<$tasks$>$ elements (see the next section for more information about tasks).
Listing \ref{lst:rng:custom} shows the RNG schema for the \textit{custom} section, and listing \ref{lst:fatigue:custom} implements the section for FATIGUE exemplarily.

\begin{itemize}
  \item \textbf{$<$recommended$>$} is used to express not binding priorities for various treatments of a specific diagnosis.
    The value of its element (ranging from 1 to 10) influences the order in which arbitrary treatments to a specific diagnoses are listed.  
  \item \textbf{$<$mandatory$>$} is used to enable care manager to enforce a specific treatment for a specific diagnoses.
\end{itemize} 

The \textit{ID} of the respective element references the ID of the recommended or mandatory task.

\lstset{label={lst:rng:custom},caption={RNG scheme of the Custom section}}
\begin{tt}
\begin{scriptsize}
\begin{lstlisting}
<element name="custom">
  <zeroOrMore>
    <element name="recommended">
      <attribute name="id">
        <text/>
      </attribute>
      <attribute name="score">
        <data type="integer">
          <param name="minInclusive">1</param>
          <param name="maxInclusive">10</param>
        </data>
      </attribute>
    </element>
  </zeroOrMore>
  <zeroOrMore>
    <element name="mandatory">
      <attribute name="id">
        <text/>
      </attribute>
    </element>
  </zeroOrMore>
</element>
\end{lstlisting}
\end{scriptsize}
\end{tt}

\lstset{label={lst:fatigue:custom},caption={Exemplary implementation of the Custom section}}
\begin{tt}
\begin{scriptsize}
\begin{lstlisting}
<custom>
  <recommended id="21" score="7"/>
  <recommended id="22" score="4"/>
  <mandatory id="30"/>
  <mandatory id="31"/>
  <mandatory id="32"/>
</custom>
\end{lstlisting}
\end{scriptsize}
\end{tt}
% }}}

\subsection{Guideline}% {{{
\label{nanda:guideline:gl}

The \textit{Guideline} section, which is defined inside the $<$guideline$>$ Tag, represents the actual care-specific knowledge and consists of the following sections:

\begin{itemize}

  \item \textbf{$<$factors$>$} contain \Nfour and \Nfive for each diagnose where they apply.
Semantically speaking, it expresses reasons that may cause symptoms or define the specific risks. 
Again, multiple $<$factor$>$ elements (with at least the attributes \textit{text} (semantic description), \textit{type} (either risk (\Nfour) or related (\Nfive)))are nested below one $<$factors$>$ element. 
The optional attributes \textit{category} is used for further refinement e.g. psychological, physiological or environmental. 
Further do $<$factor$>$ elements support additional elements including $<$hints$>$ and/or $<$examples$>$ elements (desribed above).  
It should be noted that a $<$factor$>$ element also may contain $<$factors$>$ elements but that this nesting is only supported for a depth of 1.

\item \textbf{$<$symptoms$>$} represent the building block \Nthree.
They contain multiple $<$symptom$>$ elements which itself can nest elements of the type $<$causes$>$, $<$hints$>$, and $<$examples$>$.

\item \textbf{$<$outcomes$>$} defining the evaluation criteria for treatments, which is be done everytime a treatment has ended. 
    This element nests multiple elements of the type  $<$outcome$>$. 
    $<$outcome$>$ elements themselve define three mandatory attributes, namely \textit{text} (description), \textit{source} (\Neight), and an \textit{id} (unique for the scope of the guidline). 
    Similar to the two above, an $<$outcome$>$ element nests multiple elements of the type $<$hints$>$, $<$examples$>$, and $<$inputs$>$ (which connects it to \Neleven for evaluation purposes).

\item \textbf{$<$tasks$>$} repesents sequences of activities and are therefore used to specify \Nseven, \Nnine and \Nten. 
    During the specification of them, it can be defined to execute the distinct $<$task$>$ elements, which contain the actual activities, eiter sequentially (nest them into a $<$sequential-task$>$ element) or in parallel (nest them into a $<$parallel-task$>$ element). 
    It should be noted that at this point arbitrary levels of nesting are supported, allowing to define complex tasks too. 
    We explicitely avoided supporting \textit{cycles} within guidlines as this is expressed at the (higher) level of the care plan (which is out of the scope of this technical report).
Further, both $<$sequential-tasks$>$ and $<$parallel-tasks$>$ elements support the optional attributes \textit{name} and \textit{text}. 
Each $<$task$>$ element contains the attributes \textit{id} and \textit{text}. 
%$<$labels$>$ define the corresponding NIC labels.

\end{itemize}

Additionally, each tag can contain a \textit{source} attribute, thus documenting the scientific source of the formalized knowledge. 
For example, if this attribute is applied to a tag like $<$factors$>$, the source has been used for all factors, if applied to a specific $<$factor$>$ tag, it defines that this specific factor has been formalized from this source. 
The same is true for all other tags in the \textit{Guideline} section.

The process structure defined by parallel and sequential elements in the $<$tasks$>$ element specifies the order in which certain $<$task$>$ elements are executed in a patients therapy plan. 
Additionally to the tasks, the elements from the Emphasis of Nursing Documentation section can be added to a patients therapy plan, thus emphasizing a comprehensive documentation of the patients state.

% TODO nursing documentation anhand von Stufen
    %- pflegeassessment oder neueinschätzung
    %- planung
    %- durchführung / evaluation
    %- entlassungs- oder austrittsplanung

The \textit{tasks} section of the NNN formalization as described in listing \ref{lst:rng:tasks} and \ref{lst:rng:reftask} includes all the tasks which are defined in NNN as nursing interventions.
Listing \ref{lst:fatigue:tasks} shows an excerpt of the $<$tasks$>$ section of FATIGUE. 
The implemented guideline steps have been taken from various sources, including \cite{_nanda_example_1_2012} and german literature (\cite{marilynn_e._doenges_pflegediagnosen_2002} and \cite{marlies_ehmann_pflegediagnosen_2004}). 
Listings \ref{lst:rng:documentation} and \ref{lst:rng:refdocu} define the RNG schema for the \textit{Emphases of Nursing Documentation}, as described in \cite{marilynn_e._doenges_pflegediagnosen_2002}.

\lstset{label={lst:rng:tasks},caption={RNG schema for the definition of tasks}}% {{{
\begin{tt}
\begin{scriptsize}
\begin{lstlisting}
<element name="tasks">
  <element name="labels">
    <oneOrMore>
      <element name="label">
        <attribute name="text">
          <text/>
        </attribute>
      </element>
    </oneOrMore>
  </element>
  <oneOrMore>
    <ref name="reftask"/>
  </oneOrMore>
</element>
\end{lstlisting}
\end{scriptsize}
\end{tt}% }}}

\lstset{label={lst:rng:reftask},caption={RNG definition for reftask}}% {{{
\begin{tt}
\begin{scriptsize}
\begin{lstlisting}
<define name="reftask">
  <element name="task">
    <attribute name="text">
      <text/>
    </attribute>
    <attribute name="id">
      <text/>
    </attribute>
    <optional>
      <attribute name="predictedeffort">
        <data type="integer"/>
      </attribute>
    </optional>
    <optional>
      <attribute name="score">
        <data type="integer"/>
      </attribute>
    </optional>
    <ref name="refhints"/>
    <ref name="refexamples"/>
    <ref name="refinputs"/>
  </element>
</define>
\end{lstlisting}
\end{scriptsize}
\end{tt}% }}}

\lstset{label={lst:rng:documentation},caption={RNG schema for the definition of Emphasis of Nursing Documentation}}% {{{
\begin{tt}
\begin{scriptsize}
\begin{lstlisting}
<element name="documentations">
  <oneOrMore>
    <ref name="refdocu"/>
  </oneOrMore>
</element>
\end{lstlisting}
\end{scriptsize}
\end{tt}% }}}

\lstset{label={lst:rng:refdocu},caption={RNG definition for refdocu}}% {{{
\begin{tt}
\begin{scriptsize}
\begin{lstlisting}
<define name="refdocu">
  <element name="documentations">
    <attribute name="text">
      <text/>
    </attribute>
    <attribute name="id">
      <text/>
    </attribute>
    <ref name="refhints"/>
    <ref name="refexamples"/>
    <ref name="refinputs"/>
  </element>
</define>
\end{lstlisting}
\end{scriptsize}
\end{tt}% }}}

% #####################

\lstset{label={lst:fatigue:tasks},caption={Tasks of the diagnosis FATIGUE in the NNN formalization}}
\begin{tt}
\begin{scriptsize}
\begin{lstlisting}
<tasks>
  <labels>
    <label name="Energy Management"/>
  </labels>
  <task text="Evaluate medication" id="0">
    <!-- examples can support the understanding of the related component -->
    <examples>
      <example text="Fatigue can be a byeffect of beta blockers and chemo therapy."/>
    </examples>
    <!-- Inputs can be used to define the documentation of a task -->
    <inputs>
      <input label="Short summary" xmlns="http://relaxng.org/ns/structure/1.0">
        <element name="summary">
          <data type="string">
            <param name="maxLength">50</param>
          </data>
        </element>
      </input>
      <input label="Detailed Medication" xmlns="http://relaxng.org/ns/structure/1.0">
        <element name="eingabe2">
          <text/>
        </element>
      </input>
    </inputs>
  </task>
  <task text="assess physical or psychical medical conditions" id="1">
    <examples>
      <example text="MS"/>
      <example text="Lupus"/>
      <example text="chronical pain"/>
      <example text="Hepatitis"/>
      <example text="AIDS"/>
      <example text="fear"/>
    </examples>
  </task>
  <task text="evaluate adequacy of nutrition and sleep" id="2">
    <!-- Hints can be used to give other additional information than examples -->
    <hints>
      <hint>Sometimes clients with chronic fatigue symptom can sleep excessively and need support to limit sleeping.</hint>
    </hints>
  </task>
  <task text="assess how the fatigue develops during the day" id="3">
    <inputs>
      <input label="Detailed Analysis" xmlns="http://relaxng.org/ns/structure/1.0">
        <element name="eingabe2">
          <text/>
        </element>
      </input>
    </inputs>
  </task>
  [...]
</tasks>
\end{lstlisting}
\end{scriptsize}
\end{tt}

As stated before, multiple elements necessary for implementing a comprehensive representation of the NNN guidelines cannot be represented using CPG approaches. 
Especially factors related to the diagnosis have no matching elements in the evaluated CPGs. 
Listing \ref{lst:fatigue:factors} shows how we model these building blocks.
They also contain a body which will be reused later, so we defined it separately.
This body is shown in listing \ref{lst:rng:stdbody}.

\lstset{label={lst:rng:stdbody},caption={RNG schema for the body of factors and symptoms}}% {{{
\begin{tt}
\begin{scriptsize}
\begin{lstlisting}
<define name="stdbody">
  <optional>
    <attribute name="category">
      <text/>
    </attribute>
    <optional>
      <attribute name="subcategory">
        <text/>
      </attribute>
    </optional>
  </optional>
  <attribute name="text">
    <text/>
  </attribute>
  <ref name="refhints"/>
  <ref name="refexamples"/>
</define>
\end{lstlisting}
\end{scriptsize}
\end{tt}% }}}

\lstset{label={lst:rng:factors},caption={RNG schema for the factors section}}% {{{
\begin{tt}
\begin{scriptsize}
\begin{lstlisting}
<element name="factors">
  <oneOrMore>
    <element name="factor">
      <ref name="stdbody"/>
    </element>
  </oneOrMore>
</element>
\end{lstlisting}
\end{scriptsize}
\end{tt}% }}}

\lstset{label={lst:fatigue:factors},caption={Factors of the diagnosis FATIGUE in the NNN formalization}}
\begin{tt}
\begin{scriptsize}
\begin{lstlisting}
<factors>
  <factor category="psychological" text="Boring lifestyle"/>
  <factor category="psychological" text="Stress"/>
  <factor category="psychological" text="Anxiety"/>
  <factor category="psychological" text="Depression"/>
  <factor category="environmental" text="Humidity"/>
  <factor category="environmental" text="Humidity"/>
  <factor category="environmental" text="Lights"/>
  <factor category="environmental" text="Noise"/>
  <factor category="environmental" text="Temperature"/>
  <factor category="physiologisch" text="changed chemical processes in the patient's body">
    <factors>
      <factor text="medicines"/>
      <factor text="drug withdrawal"/>
      <factor text="chemotherapy"/>
    </factors>
    <hints>
      <hint text="e.g. because of medicines, drug withdrawal or other reasons"/>
    </hints>
  </factor>
  [...]
</factors>
\end{lstlisting}
\end{scriptsize}
\end{tt}

The defining characteristics are possible symptoms a patient can show when he suffers the given diagnosis. 
Listing \ref{lst:fatigue:symptoms} shows how symptoms can be modeled in our formalization.

\lstset{label={lst:rng:symptoms},caption={RNG schema for the symptoms section}}% {{{
\begin{tt}
\begin{scriptsize}
\begin{lstlisting}
<element name="symptoms">
  <oneOrMore>
    <element name="symptom">
      <ref name="stdbody"/>
    </element>
  </oneOrMore>
</element>
\end{lstlisting}
\end{scriptsize}
\end{tt}% }}}

\lstset{label={lst:fatigue:symptoms},caption={Symptoms of the diagnosis FATIGUE in the NNN formalization}}
\begin{tt}
\begin{scriptsize}
\begin{lstlisting}
<symptoms>
  <symptom category="subjective" text="inability to restore energy even after sleep"/>
  <symptom category="subjective" text="lack of energy or inability to maintain usual level of physical activity"/>
  <symptom category="subjective" text="increase in the rest requirements"/>
  <symptom category="subjective" text="tired"/>
  <symptom category="subjective" text="lethargic"/>
  <symptom category="subjective" text="listless"/>
  <symptom category="subjective" text="perceived need for additional energy to accomplish routine tasks"/>
  <symptom category="subjective" text="introspection"/>
  <symptom category="subjective" text="compromised libido"/>
  <symptom category="subjective" text="feeling of guilt for not keeping up with responsibilities"/>
  <symptom category="subjective/objective" text="inability to maintain usual routines"/>
  <symptom category="subjective/objective" text="compromised concentration"/>
  <symptom category="subjective/objective" text="disinterest in surroundings"/>
  <symptom category="subjective/objective" text="drowsy"/>
  <symptom category="objective" text="decreased performance"/>
  <symptom category="objective" text="increase in physical complaints"/>
  <symptom category="objective" text="verbalization of an unremitting and overwhelming lack of energy"/>
</symptoms>
\end{lstlisting}
\end{scriptsize}
\end{tt}

Outcomes, as defined in the Nursing Outcomes Classification, are seamlessly integrated into our formalization. 
Listing \ref{lst:fatigue:outcome} shows three exemplary outcomes and the respective NOC labels. 
While the first two outcomes implement documentationary $<$input$>$ items for assessing whether a patient has reached the goal or not, the last outcome includes the possibility of documenting the way a patient describes his plan of conserving energy.

\lstset{label={lst:rng:outcome},caption={RNG schema for the outcome section}}% {{{
\begin{tt}
\begin{scriptsize}
\begin{lstlisting}
<element name="outcomes">
  <element name="labels">
    <oneOrMore>
      <element name="label">
        <attribute name="text">
          <text/>
        </attribute>
      </element>
    </oneOrMore>
  </element>
  <oneOrMore>
    <element name="outcome">
      <attribute name="goal">
        <choice>
          <value>achieve</value>
          <value>maintain</value>
          <value>prevent</value>
        </choice>
      </attribute>
      <attribute name="text">
        <text/>
      </attribute>
      <attribute name="id">
        <text/>
      </attribute>
      <ref name="refhints"/>
      <ref name="refexamples"/>
      <ref name="refinputs"/>
    </element>
  </oneOrMore>
</element>
\end{lstlisting}
\end{scriptsize}
\end{tt}% }}}

\lstset{label={lst:fatigue:outcome},caption={Exemplary outcomes for FATIGUE}}
\begin{tt}
\begin{scriptsize}
\begin{lstlisting}
<outcomes>
  <labels>
    <label text="Endurance"/>
    <label text="Concentration"/>
    <label text="Energy Conservation"/>
    <label text="Nutrition Status: Energy"/>
  </labels>
  <outcome goal="achieve" text="The patient verbalizes increased energy.">
    <inputs>
      <input label="Goal reached" xmlns="http://relaxng.org/ns/structure/1.0">
        <element name="select">
          <element name="yes">Yes</element>
          <element name="no">No</element>
        </element>
      </input>
    </inputs>
  </outcome>
  <outcome goal="achieve" text="The patient verbalizes improved well-being.">
    <inputs>
      <input label="Goal reached" xmlns="http://relaxng.org/ns/structure/1.0">
        <element name="select">
          <element name="yes">Yes</element>
          <element name="no">No</element>
        </element>
      </input>
    </inputs>
  </outcome>
  <outcome goal="achieve" text="The patient explains energy conservation plan to offset fatigue."/>
  <inputs>
    <input label="Patients explanation" xmlns="http://relaxng.org/ns/structure/1.0">
      <element name="input-pat">
        <text/>
      </element>
    </input>
  </inputs>
</outcome>
[...]
</outcomes>
\end{lstlisting}
\end{scriptsize}
\end{tt}

\section{Discussion}% {{{
\label{ref:discussion}

The NNN formalization presented in this paper constitutes an initial step into the field of 'computer-aided' nursing in the care domain as envisioned by the ACaPlan project by providing the basic building blocks required for comprehensive and formalized care planning and execution. 

During development of the NNN formalization, the discrepancies between the medical and the care domain based on the differences between respective guidelines as described in chapter \ref{ref:evaluation} became obvious very quickly, although these fields of research are related on many levels.
The fact, that the revised CIGs do not support any means for implementing risks and related factors resulted in the need for developing an approach to formalize knowledge specific for NNN guidelines. 
The implementation of NNN guidelines in this technical report covers all relevant parts of a diagnosis, so a patient's state can be assessed from various points of view: 
On the one hand, based on data about his history, risks and related factors can be used to diagnose potential threats very early; on the other hand it is possible to use symptoms a patient shows to determine his diagnoses. 
These different perspectives support nursing personnel when creating therapy plans in many situations. 
Additionally, by introducing $<$input$>$ tags, different scales and documentationary forms can be added to support the documentation.

Evaluations with domain experts show that the formalization presented in this technical report addresses all relevant parts of NANDA, NIC and NOC. 
Thus, guidelines formalized using this approach can be applied for supporting nurses in real life scenarios.
% }}}

\noindent{\bf Acknowledgment:} We thank Adelheid Beyerl and Martin Zigler from the care centers St. P\"olten Pottenbrunn and Vitacon for the discussions and insights into the care domain as well as the great cooperation within the ACaPlan project. 

\bibliographystyle{IEEEtran}
\bibliography{output}

\end{document}